\documentclass[runningheads]{llncs}
\usepackage[T1]{fontenc}
\usepackage{graphicx,verbatim}
\usepackage{amsmath}
\usepackage{amssymb}
\begin{document}
\title{Scaling Ultrasound Volumetric Reconstruction via Mobile Augmented Reality}
\titlerunning{Volumetric Ultrasound with Mobile Augmented Reality}
%
\author{Kian Wei Ng\inst{1,2} \and
Yujia Gao\inst{1} \and
Deborah Khoo\inst{1} \and
Ying Zhen Tan\inst{1,2} \and
Chengzheng Mao\inst{1} \and
Haojie Cheng\inst{2} \and
Andrew Makmur\inst{1} \and
Kee Yuan Ngiam\inst{1} \and
Serene Goh\thanks{Last Author}\inst{1} \and
Eng Tat Khoo\thanks{Corresponding Author}\inst{2}
}
\authorrunning{K.W. Ng et al.}
\institute{National University Health System, Singapore\and
College of Design and Engineering, National University of Singapore, Singapore}

  
\maketitle              
\begin{abstract}
Accurate volumetric characterization of lesions is essential for oncologic diagnosis, risk stratification, and treatment planning. While imaging modalities such as Computed Tomography provide high-quality 3D data, 2D ultrasound (2D-US) remains the preferred first-line modality for breast and thyroid imaging due to cost, portability, and safety factors. However, volume estimates derived from 2D-US suffer from high inter-user variability even among experienced clinicians. Existing 3D ultrasound (3D-US) solutions use specialized probes or external tracking hardware, but such configurations increase costs and diminish portability, constraining widespread clinical use. To address these limitations, we present Mobile Augmented Reality Volumetric Ultrasound (MARVUS), a resource-efficient system designed to increase accessibility to accurate and reproducible volumetric assessment. MARVUS is interoperable with conventional ultrasound (US) systems, using a foundation model to enhance cross-specialty generalization while minimizing hardware requirements relative to current 3D-US solutions. In a user study involving experienced clinicians performing measurements on breast phantoms, MARVUS yielded a substantial improvement in volume estimation accuracy (mean difference: 0.469 cm³) with reduced inter-user variability (mean difference: 0.417 cm³). Additionally, we prove that augmented reality (AR) visualizations enhance objective performance metrics and clinician-reported usability. Collectively, our findings suggests that MARVUS can enhance US-based cancer screening, diagnostic workflows, and treatment planning in a scalable, cost-conscious, and resource-efficient manner.


\end{abstract}

\section{Introduction}\label{sec:introduction}
Volumetric assessment of suspicious nodules in the breast and thyroid inform clinical decision making. In routine screening, volume estimates help to quantify nodule growth, informing downstream treatment \cite{cite_thyata,cite_breastsubjmorph}. In treatment planning, estimates influence key intervention choices, impacting cost and quality of life outcomes \cite{cite_siemensuseful}. In surgery, knowledge of the tumor’s 3D structure enables precise resections, minimizing healthy tissue loss while ensuring oncological safety \cite{cite_breastnodulemesh}. 

US offers advantages over Magnetic Resonance Imaging, with lower costs, minimal infrastructure, real-time imaging, and higher portability \cite{cite_modalitycomp}. However, volumetric assessment with standard 2D-US handheld probes is challenging. Interpreting 3D structures from 2D images lead to high inter-operator variability, with up to 48.96\% variation in nodule volume estimates \cite{cite_thyinterobs}. This variability also stems from manual volume estimation formulas, as nodules rarely match ideal spherical assumptions. Volume aside, subjective interpretation of lesion morphology can lead to inconsistent clinical decisions \cite{cite_breastsubjmorph}. As accurate and consistent tumor size measurements form the basis for clinical staging and treatment planning \cite{cite_breastenhancebenefits}, 3D-US solutions have been proposed to address existing limitations.

\begin{table*}[t]
\caption{System comparison for related 3D-US and Augmented Reality works\label{tab:table1}}
\centering
\begin{tabular}{lcccc}
\hline
\textbf{} & \textbf{Ultrasound} & \textbf{Tracking} & \textbf{Augmented} & \textbf{Segmentation} \\
\textbf{Ref.} & \textbf{Probe} & \textbf{Sensor} & \textbf{Reality Device} & \textbf{Algorithm} \\
\hline
\textbf{\cite{cite_robotickuka}} & Standard (2D) & Robotic Arm & - & Specialized\\
\textbf{\cite{cite_realsensesegmrecon}} & Standard (2D) & Depth Camera & - & Specialized\\
\textbf{\cite{cite_arreconspec}} & Standard (2D) & Stereo Camera & Headset & Manual\\
\textbf{\cite{cite_ar3dprobecalibbiop}} & Specialized (3D) & Stereo Camera & Headset & Specialized\\
\hline
\textbf{Ours} & Standard (2D) & Phone & Phone & Prompt-based\\
\hline
\end{tabular}
\end{table*}

\textbf{3D-US Acquisition Systems --} 3D-US volumetric data can be directly acquired through dedicated 3D transducers \cite{cite_siemensuseful,cite_2dvs3dusbreast}. Aside from specialized 3D-US setups, external tracking systems (e.g. robotic arm \cite{cite_robotickuka}, stereo cameras \cite{cite_arreconspec,cite_polarisrecon}, depth cameras \cite{cite_realsensesegmrecon}) can be used to compound 3D-US data from standard 2D-US images. Spatial alignment between the external tracking system and US probe is done through a calibration process involving a calibration phantom \cite{cite_calibreprod2008,cite_ar3dprobecalibbiop} or customized tracking attachment \cite{cite_arreconspec}.

\textbf{Segmentation --} To process the densely acquired 3D-US data into clinically relevant signals (e.g. volume estimation in cm³, or morphological features as a mesh), segmentation has to be performed. Typically, specialized segmentation models are trained to extract and reconstruct the target nodule structure \cite{cite_realsensesegmrecon,cite_ar3dprobecalibbiop,cite_arreconspec}. 
These nodule structures can then be used to derive clinically relevant properties such as estimated volume growth, or morphological features \cite{cite_thyata,cite_birads}.

\textbf{Augmented Reality 3D-US Visualizations --} Augmented reality has been used to overlaying preoperative CT or MRI scans onto the patient's body to enhance spatial understanding \cite{cite_arnav,cite_mrnavtr,cite_mrnav}. Extending into US, \cite{cite_arreconspec} used an AR headset (Microsoft HoloLens) to track a standard 2D-US probe (Mindray DC-80A) through a custom attachment. Post manual segmentation, the resulting 3D structure was rendered in AR for the purpose of biopsy guidance. \cite{cite_ar3dprobecalibbiop} followed a similar workflow, but opted for a specialized 3D-US probe (Philips xMatrix) for data acquisition, with a reported setup and processing time of 20-25 minutes.

The existing 3D-US reconstruction and volume estimation pipelines (summarized in Table 1) are ill suited for scaling across clinical specialties, and are also too demanding for low-resource clinical settings. To address this, we propose a Mobile Augmented Reality Volumetric Ultrasound (MARVUS) system using standard mobile devices. Reduced hardware requirements and simplifications in workflow improve scalability and efficiency, with AR visualizations improving performance and operator-system trust \cite{cite_barriertrustdata}. Our contributions are as follows:

\begin{enumerate}
\item{\textbf{Scaling Low-cost 3D-US:} We propose a mobile-based 3D-US system with a novel calibration setup that scales across US probes and clinical specialties, using a foundation model to enable easy cross-specialty deployment.}
\item{\textbf{AR-Enhanced Workflow:} We introduce an AR visualizations that guide operators through volumetric data acquisition, enhancing measurement accuracy and operator confidence through an intuitive real-time feedback.}
\item{\textbf{Expert Validation:} Evaluating MARVUS with experienced clinicians and realistic, clinically derived phantoms showed significant improvements in volumetric accuracy and reduction in inter-operator variability.}
\end{enumerate}

\section{Methods}

\begin{figure*}[!t]
\centering
\includegraphics[width=0.85\columnwidth]{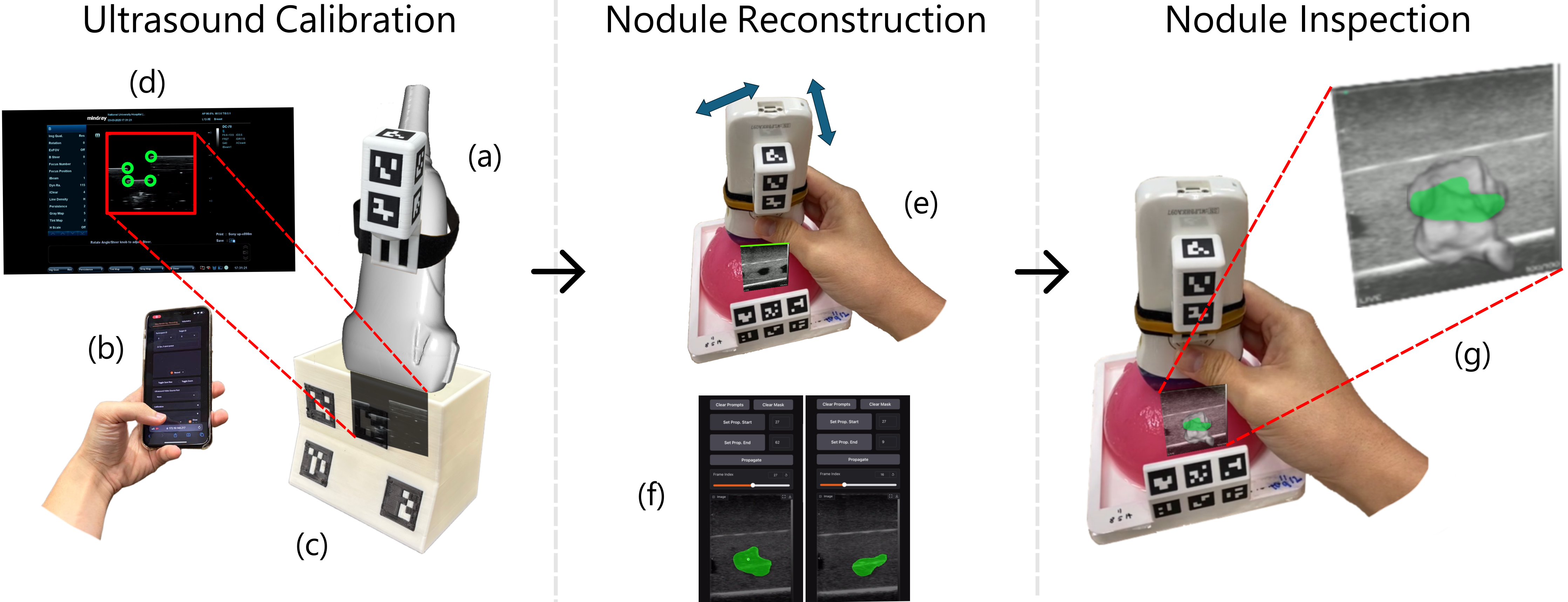}
\caption{The MARVUS workflow. \textbf{Calibration --} US probe with strapped ArUco markers (a) is tracked via a mobile phone (b). Calibration is done via a simple phantom (c), with region-of-interest and keypoints detected from a raw screen capture (d). \textbf{Reconstruction --} A freehand sweep imaging a nodule is done with accompanying augmented reality visualizations (e). Nodule segmentation is obtained in a semi-automated manner (f) and compounded into 3D for volume estimation. \textbf{Verification --} The 3D mesh is verified in augmented reality (g) by comparing the mesh-plane intersection with a live ultrasound feed (green denotes intersection between the mesh and the live US image).}
\label{fig1_main}
\end{figure*}

We describe our workflow (Fig. 1), including calibration, followed by nodule assessment (example video available online https://youtu.be/m4llYcZpqmM) via acquisition, reconstruction, and verification stages.

\subsection{Ultrasound Calibration (Intrinsics)}

Raw images from US machines via video capture cards contain redundant rendered UI elements. A "Region-of-Interest" (ROI) is first computed, defining a fixed area updated with live US sensor data. A set of $n$ US images $M_{\text{US}} \in \mathbb{R}^{n \times h \times w}$ is processed pixel-wise for the deviation of each pixel over time. After percentile-based outlier rejection to account for spurious changes (e.g. cursor movement), a binary activity map $M_{activity} \in \mathbb{R}^{h \times w}$ can be computed. Finally, a ROI box $(x_{roi}, y_{roi}, w_{roi}, h_{roi})$ is fitted on large activity regions, ignoring small regions corresponding to consistent UI updates (e.g. rendered timestamps). 

The next stage of calibration involves solving for the US image scale $s_{us}$ in $mm/pixel$. $s_{us}$ is typically determined via a calibration phantom containing fiducial structures that are visible in US images. Instead of standard "n-wire" phantom variants comprising multiple wires secured in a water bath \cite{cite_firstwire,cite_calibreprod2008}, we introduce a novel single-part US calibration phantom (Fig. 2) that enables single frame calibration. This leads to a simpler fabrication and setup process, while also reducing the calibration time from several minutes \cite{cite_firstwire} to seconds.

\begin{figure}[t]
\centering
\includegraphics[width=0.84\columnwidth]{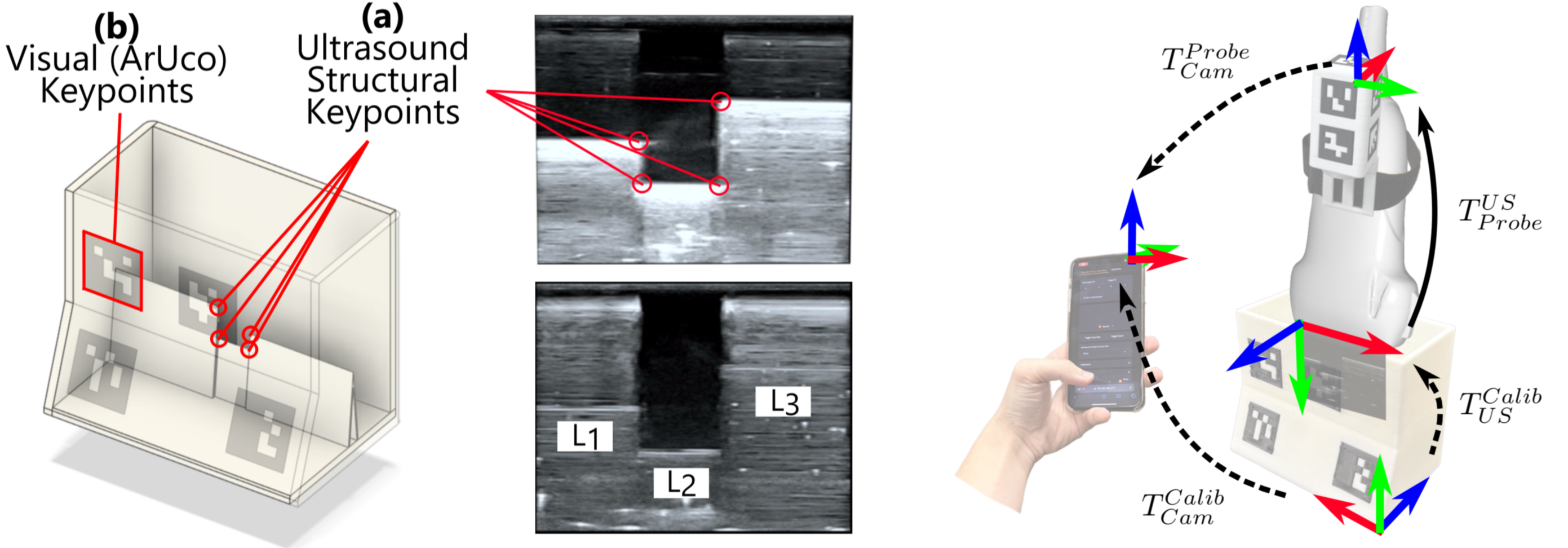}
\caption{\textbf{US Calibration Phantom}. A single-piece 3D-printed phantom is used for single-frame spatial calibration using US-visible "ledges" \textbf{(a)} with intersections enabling calibration. The "ledges" $(L_{1},L_{2},L_{3})$ are brightest when viewed directly in-plane due to US physics. \textbf{(b)} Visible markers are used for extrinsics calibration (right).}
\label{fig3_calibscheme}
\end{figure}

For each US frame acquired with the probe in the phantom, we detect "ledge"-based keypoints (Fig. 2(a)). A filter kernel and Hough transform enhance horizontal linear features and extract candidate lines \cite{cite_structuringelement,cite_hough}. These are searched to find line triplets $(L_{1},L_{2},L_{3})$ that satisfies the known geometry, yielding four keypoints per frame (Fig. 2(a)) and a scale estimate. Median pooling across frames gives $\hat{s_{us}}$. Finally, using the 0.5 cm depth increments typical of US systems, we refine $s_{us}$ such that $s_{us} \times h \in \{5, 10, ... 100\}$.

This cumulates in $K_{us}$, which allows for pixels to be represented in 3D:

\begin{align}
K_{us}&=
\begin{bmatrix} 
s_{us} & 0 & 0 & -x_{roi}*s_{sus}\\
0 & s_{us} & 0 & -y_{roi}*s_{sus}\\
0 & 0 & 1 & 0\\
0 & 0 & 0 & 1
\end{bmatrix} && \begin{bmatrix}
x_{us}\\y_{us}\\z_{us}\\1
\end{bmatrix} =
K_{us} *
\begin{bmatrix}
i_{us}\\j_{us}\\0\\1
\end{bmatrix}
\end{align} 

\subsection{Ultrasound Calibration (Extrinsics)}

Like other 2D-US retrofits for 3D reconstruction \cite{cite_robotickuka,cite_ar3dprobecalibbiop,cite_arreconspec}, MARVUS requires external calibration. The markers on the probe are tracked through the live camera feed, attaining a time-varying $T^{Probe}_{Cam}$. To accurately reflect the US image position along the transducer tip, a fixed offset $T^{US}_{Probe}$ needs to be applied. $T^{US}_{Probe}$ is fixed, requiring recomputation only when the strap is adjusted to suit a new user's ergonomic preference. $T^{US}_{Probe}$ is computed by linking several transforms:
\begin{align}
    T^{US}_{Probe} = {(T^{Calib}_{US})}^{-1} * T^{Calib}_{Cam} * {(T^{Probe}_{Cam})}^{-1}
\end{align}

$T^{Calib}_{US}$ is computed by tracking the location of the calibration phantom in the US image. The four distinct keypoints from the previous stage (Fig. 2(a)) are projected to 3D space using $K_{us}$ (Eq. 1), and used to compute the pose transformation matrix $T^{Calib}_{US}$ through a least square Kabsch algorithm \cite{cite_alignvectors}.

To complete the transformation chain (Eq. 2), $T^{Calib}_{Cam}$ and $T^{Probe}_{Cam}$ are obtained by tracking ArUco marker sets (Fig. 2b) \cite{cite_aruco}. Unlike prior planar marker setups \cite{cite_planartrack,cite_arreconspec}, we use a non-planar configuration for improved tracking \cite{cite_planarbad}, with 4 markers on the calibration phantom (Fig. 3b) and 9 on the probe tracker (Fig. 1a). Monocular pose estimation uses corner pixel detection \cite{cite_aruco}, Perspective-n-Point \cite{cite_pnp}, with randomrandom sampling and consensus \cite{cite_ransac}, and 
a spatio-temporal 1€ filter for noise suppression \cite{cite_ransac,cite_oneeurofilter}.

\subsection{Nodule Reconstruction}

\begin{figure}[t]
\centering
\includegraphics[width=0.88\columnwidth]{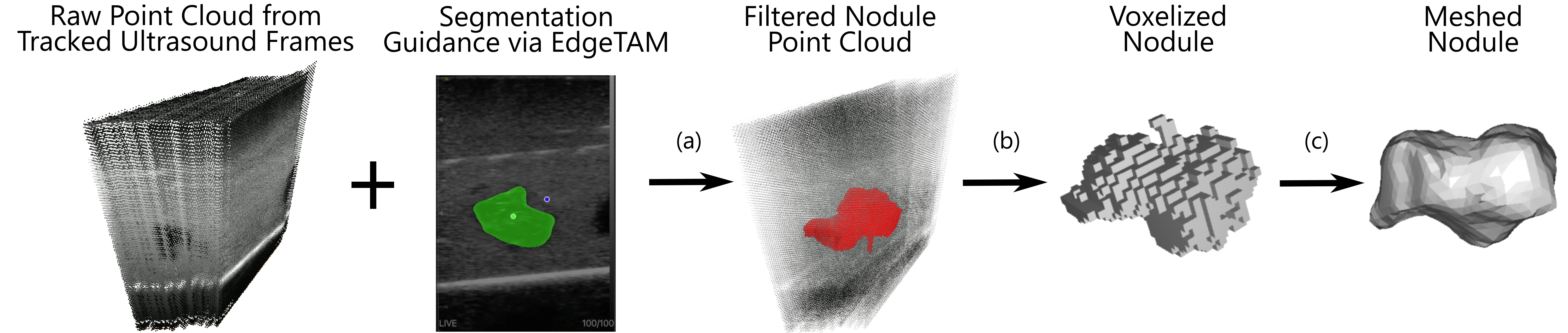}
\caption{\textbf{Nodule Reconstruction Steps}. \textbf{(a)} Segmentation of tracked US sequence produces a point cloud. \textbf{(b)} Voxelization resamples and fills sparsely sampled locations. \textbf{(c)} A meshing algorithm is applied to smoothen noise-induced highly irregular surfaces.}
\label{fig5_meshing}
\end{figure}

Reconstruction of US image sequences requires a static spatial reference frame $T^{Ref}_{Cam}$, achieved via another ArUco marker set (Fig. 1e). In practice, placing the marker set on the patient allows free camera motion while correcting for coarse patient motion. $T^{Ref}_{Cam}$ is tracked similar to $T^{Probe}_{Cam}$, and used as following:

\begin{align}
    T^{US}_{Ref} = T^{US}_{Cam} * (T^{Ref}_{Cam})^{-1}
\end{align}

By computing and applying $T^{US}_{Ref}$ to consecutive US frames, a textured point cloud containing 3D-US data is formed (Fig. 3, left). The dense point cloud is filtered via segmentation for the target nodule. While specialized deep learning models are often used for this task \cite{cite_robotickuka,cite_realsensesegmrecon}, such models suffer from high data and annotation costs, presenting a barrier to adoption across institutes and specialties \cite{cite_barriertrustdata}. We use EdgeTAM as a generalized semi-automated approach \cite{cite_edgetam}, where masks are generated via "point" user prompts (Fig. 3, centre), and multi-frame video segmentation is efficiently done via automated propagation. The multi-frame segmentation allows the dense point cloud to be reduced into one that contains only the target nodule (Fig. 5a).

Because the point cloud is generated from a free-hand sweep, point density is not uniform. Sparsely sampled regions from faster sweeping motion are addressed through a spatial resampling voxelization step (at $1$mm$^3$) (Fig. 5b). Lastly, the voxel grid is turned into a mesh via marching cubes \cite{cite_vtk}, which smooths the surface to remove outliers caused by pose estimation noise (Fig. 5c). Furthermore, a closed mesh representation can be further processed to extract an estimation for the nodule volume, useful for downstream clinical decision making.

\subsection{Nodule Assessment (Augmented Reality)}

With the nodule reconstructed, the operator can inspect its topology in isolation using a standard mesh viewer (Fig. 5c). This is enhanced through an AR implementation with two modes to improve operator trust and system acceptability.

The mesh is projected at its original location beneath the surface (Fig. 1g), enabling assessment of the nodule relative to surface landmarks. Beyond a static overlay, intersection of the reconstructed mesh with live US images is shown (Fig. 1g, green). Computed via signed-distance field, this enables scrutiny of reconstruction and localization quality: an error-free reconstruction yields perfect overlap between the green reconstruction slice and live US nodule images.

\subsection{User Study Design}

\begin{table}[t]
\caption{User Study: Volume Assessment Methods\label{tab:table2}}
\centering
\begin{tabular}{lccc}
\hline
\textbf{} & \textbf{Image Acquisition } & \textbf{ Volume Estimation } & \textbf{ Nodule Assessment}\\
\hline
\textbf{Control} & Manual & Ellipsoid Formula & N/A\\
\textbf{Recon} & Sweep & Automated & Mesh\\
\textbf{Recon+AR} & Sweep (AR) & Automated & Mesh (AR)\\
\hline
\end{tabular}
\end{table}

We design a user study to test our hypotheses that \textbf{1)} MARVUS improves volume estimation accuracy, and \textbf{2)} our AR visuals improve system usability.

Nodules were sampled from the MAMA-MIA dataset \cite{cite_breastnodulemesh} to reflect real-world nodule complexity. Twelve nodules were scaled to a fixed volume ($1.69cm^3$) and sorted into groups (\textbf{"Control", "Recon", "Recon+AR"}), ensuring balanced complexity (via sphericity $\psi_{GT} \in [0,1]$ \cite{cite_sphericity}). Casting in a 3D printed mold with konjac, coloring and flour resulted in opaque, anechoic target nodules \cite{cite_konnyaku}.

We recruited eight experts (>5 years US experience) for the study. Participants used the three methods (summarized Table II) to estimate nodule volumes, with order randomized to negate sequence bias. \textbf{"Control" --} participants followed standard clinical practice, using two orthogonal planes to estimate volume via an ellipsoid assumption \cite{cite_breastellipsoid}. \textbf{"Recon" --} participants performed a sweep across the nodule, using EdgeTAM to segment the nodule from the US video. After reconstruction, participants assessed the nodule structure in isolation (similar to final visual in Fig. 3c). \textbf{"Recon+AR" --} participants followed the same process as in "Recon", with added AR visuals in the sweep acquisition (Fig. 1e) and post-reconstruction (Fig. 1g) steps. A example of \textbf{"Recon+AR"} MARVUS usage is available online (https://youtu.be/m4llYcZpqmM). Upon task completion, NASA-TLX and SUS surveys were used to assess subjective task load and system usability across volume estimation methods.

\section{Results}

\begin{figure}[t]
\centering
\includegraphics[width=.75\columnwidth]{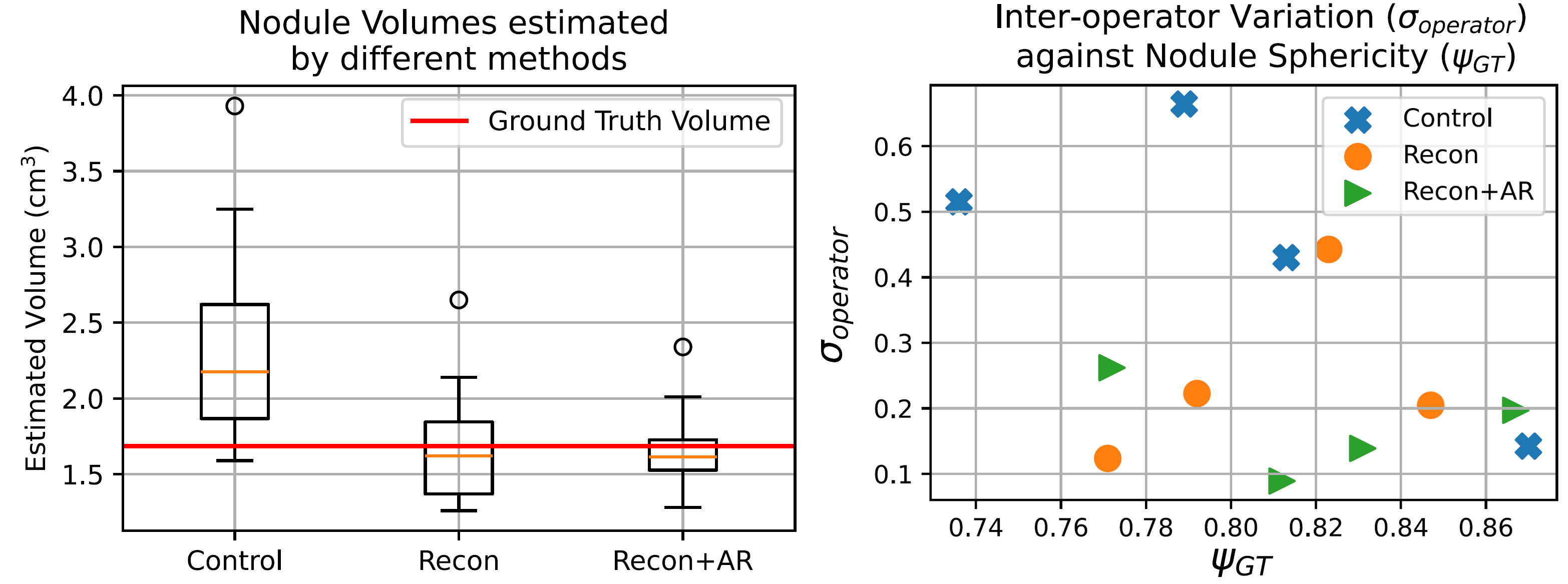}
\caption{\textbf{Nodule Volume Estimation}. Our AR 3D-US setup improves accuracy and reduces operator variability \textbf{(left)}, notably for complex non-spherical nodules \textbf{(right)}.}
\label{fig7_resmaingroup}
\end{figure}

\begin{table}[t]
\centering
\caption{US Calibration: Calibration Reproducibility (CR) against Related Works\label{tab:table5}}
\begin{tabular}{lcc}
\hline
\textbf{Tracker} & \textbf{Phantom Type} & \textbf{CR (mm)} \\
\hline
Optotrak Certus (NDI) \cite{cite_calibreprod2008} & Double-N Wire & $0.938\pm0.118$\\
Vicra (NDI) \cite{cite_calibreprod2018} & Arbitrary Wire & $0.896\pm0.075$\\
Polaris Spectra (NDI) \cite{cite_calibreprod2013} & Triple-N Wire & $0.690\pm0.310$ \\
\hline
Mobile Phone (Ours) & 3D-printed Ledge & $0.826\pm0.447$ \\
\hline
\end{tabular}
\end{table}

\subsection{Ultrasound Calibration}

We evaluate our proposed calibration phantom setup (Fig. 3) on a standard US system (Samsung LA3-14AD). Generalization is assessed by varying imaging depth (30, 40, 50 mm) and gain (low, high gain). Each unique parameter combination is calibrated 5 times, with 30 calibration runs in total.

We use the Calibration Reproducibility (CR) metric introduced in \cite{cite_calibreprod2008} to assess overall calibration quality. The overall CR error for our proposed calibration system is $0.826\pm0.447$ mm, ranging from $0.515$ to $1.005$ mm across different imaging settings. This makes it consistent and competitive with existing works (Table III), without the need for specialized surgical grade optical sensors and time-consuming wire-based calibration processes \cite{cite_calibreprod2008,cite_calibreprod2018,cite_calibreprod2013}.

\subsection{User Study (Volumetry)}

Our 3D-US setup \textbf{("Recon")} resulted in significant reduction (wilcoxon rank-sum test, $p<0.05$) of volume error and inter-operator variability ($0.270\pm0.189cm^3$) compared to baseline \textbf{("Control")} ($0.630\pm0.549 cm^3$). This mirrors similar works \cite{cite_siemensuseful,cite_2dvs3dusbreast}, indicating that our setup retains the advantage of 3D-US, while introducing cost and scalability benefits. The proposed addition of AR \textbf{("Recon+AR")} further reduced volume error and variability ($0.161\pm0.132cm^3$) (Fig. 4, left). This suggests that our AR visuals provides a spatially intuitive interface that improves probe handling and tracking.

Using the inverse of sphericity $\psi_{GT}$ as a proxy for nodule complexity (Fig. 4, right), we note that nodules that undergo baseline volume estimation \textbf{("Control")} have higher inter-operator variability when complexity increases ($\downarrow\psi_{GT}$). In comparison, nodule volumes estimated via \textbf{"Recon"} and \textbf{"Recon+AR"} exhibit more consistent performance across $\psi_{GT}$ ranges. This suggests that examination of structurally complex nodules could benefit from MARVUS by removing manual US plane selection and measurement subjectiveness.

\subsection{User Study (Survey Response)}

The NASA-TLX responses indicates no significant difference between methods (wilcoxon signed-rank test, $p>0.05$) \cite{cite_nasatlx}. Perceived effort and physical task load was lower for \textbf{"Recon"}/\textbf{"Recon+AR"}, stemming from the switch from a manual search to a simple data acquisition sweep. Conversely, factors such as mental load and frustration was lowest for \textbf{"Control"}, likely due to our participants' high level of clinical experience, where deviation from a familiar setup introduces friction. Nonetheless, aggregate task load scores showed a slight improvement when using \textbf{"Recon"}/\textbf{"Recon+AR"}, indicating non-inferiority and potential benefits even for experienced operators.

Finally, we used the SUS scale to assess the impact of AR visuals on perceived system usability and adoption \cite{cite_sus}. With a $100$ point total, system usability was reported at $65.625\pm11.575$ without AR \textbf{("Recon")}. This improved to $68.438\pm14.249$ with AR visuals \textbf{("Recon+AR")}. While overall scores were not significantly different (wilcoxon signed-rank test, $p>0.05$), the change in response to \textit{“I felt very confident using the system”} was statistically significant (wilcoxon signed-rank test, $p<0.05$), improving from $1.500\pm0.707$ without AR \textbf{("Recon")} to $2.375\pm0.484$ when AR was used \textbf{("Recon+AR")}. This suggests that the proposed addition of post-reconstruction AR visuals (Fig. 1g) improved transparency into the reconstruction results, increasing operator confidence by presenting results in a spatially intuitive and clear manner.

\section{Conclusion}

This paper advances the intersection between 3D-US and AR, with the design, implementation and phantom study validation of an accessible and efficient system, using standard (2D-US probe, mobile device) and simplified (calibration phantom) components. With significant results in improving volume estimation performance (higher accuracy, lower user variability), follow up works will investigate the potential of our system in more diverse clinical scenarios and specialties, including on nodules of varying sonographic characteristics. Further applications will be explored, including in biopsy guidance, or automated and AR-enhanced surgical residency training made possible by the low system cost.

%
\bibliographystyle{splncs04}
\bibliography{references}

\end{document}